\documentclass[]{spie}  

\usepackage{amsmath,amsfonts,amssymb}
\usepackage{graphicx}
\usepackage[colorlinks=true, allcolors=blue]{hyperref}
\usepackage{booktabs}
\usepackage{multirow}

\title{AdverX-Ray: Ensuring X-Ray Integrity Through Frequency-Sensitive Adversarial VAEs}

\author[a]{Francisco Caetano$^\dagger$}
\author[a]{Christiaan Viviers$^\dagger$}
\author[b]{Lena Filatova}
\author[a]{\\Peter H.N. de With}
\author[a]{Fons van der Sommen}
\affil[a]{Eindhoven University of Technology, Groene Loper 3, 5612 AE, Eindhoven, Netherlands}
\affil[b]{Philips IGT, 5684 PC, Best, Netherlands}

\authorinfo{$^\dagger$These authors contributed equally to this work.\\Further author information and correspondence to F.C. E-mail: f.t.de.espirito.santo.e.caetano@tue.nl}

\begin{document} 
\maketitle

\begin{abstract}
Ensuring the quality and integrity of medical images is crucial for maintaining diagnostic accuracy in deep learning-based Computer-Aided Diagnosis and Computer-Aided Detection~(CAD) systems. Covariate shifts are subtle variations in the data distribution caused by different imaging devices or settings and can severely degrade model performance, similar to the effects of adversarial attacks. Therefore, it is vital to have a lightweight and fast method to assess the quality of these images prior to using CAD models. AdverX-Ray addresses this need by serving as an image-quality assessment layer, designed to detect covariate shifts effectively. This Adversarial Variational Autoencoder prioritizes the discriminator's role, using the suboptimal outputs of the generator as negative samples to fine-tune the discriminator's ability to identify high-frequency artifacts. Images generated by adversarial networks often exhibit severe high-frequency artifacts, guiding the discriminator to focus excessively on these components. This makes the discriminator ideal for this approach. Trained on patches from X-ray images of specific machine models, AdverX-Ray can evaluate whether a scan matches the training distribution, or if a scan from the same machine is captured under different settings. Extensive comparisons with various OOD detection methods show that AdverX-Ray significantly outperforms existing techniques, achieving a 96.2-\% average AUROC using only 64 random patches from an X-ray. Its lightweight and fast architecture makes it suitable for real-time applications, enhancing the reliability of medical imaging systems. The code and pretrained models are publicly available\footnote{\url{https://github.com/caetas/AdverX}}.
\end{abstract}

\section{INTRODUCTION}

The accurate and reliable detection of out-of-distribution~(OOD) data is paramount to diagnostic accuracy and overall reliability of medical imaging systems, ultimately ensuring patient safety. Faulty systems processing incorrect imaging statistics can adversely affect diagnostics. It is well established that modern deep learning-based Computer-Aided Diagnosis and Computer-Aided Detection~(CAD) systems are vulnerable to distribution shifts, which can lead to erroneous predictions.

Conventional anomaly detection techniques often rely on identifying deviations from a learned statistical representation of the in-distribution~(ID) data and typically focus on semantic anomalies in images~\cite{yang2024generalized}. Medical images can exhibit complex noise patterns and variability due to equipment variations, imaging conditions, and physiological differences between patients~\cite{goyal2018noise_issues}. Previous research on the impact of covariate shifts on X-ray chest images has demonstrated that factors such as acquisition parameters, device manufacturers, and geographical variations can degrade the $F_1$ score by up to 6\%~\cite{bercean2023breaking}. Identifying these covariate factors, i.e. the change in the distribution of high-level image statistics~(covariates) subject to consistent low-level semantics~\cite{tian2021exploringcovariate}, will enable safer deployment and continuous exploitation of modern medical imaging systems, while ensuring that CAD methods remain applied within their ID range.

Detecting covariate factors or sensory anomalies of OOD is challenging due to the limited availability of OOD samples. Consequently, unsupervised OOD detection methods are particularly attractive, since they leverage only ID data to construct a representation of normal samples. This representation can then be used to evaluate the deviation of new test samples from the learned model.

This work employs various unsupervised generative models to detect OOD covariate shifts in X-ray images. The methods are evaluated on two X-ray imaging datasets: (1)~a newly collected dataset featuring different imaging settings as OOD sets, and (2)~the BIMCV-COVID19+ dataset~\cite{vaya2020bimcv}, which includes chest X-ray~(CR, DX) and computed tomography~(CT) images of COVID-19 patients from various systems. The limited performance of existing methods on the proposed task prompted the development of a novel unsupervised method, called AdverX-Ray. The research objective is to exploit AdverX-Ray to establish a high-quality prediction of the correct operation of the X-ray system and its corresponding output image quality.

Within adversarial environments, Batch Normalization~(BN)~\cite{ioffe2015batch} has been noted for its inherent ability to differentiate between real~(ID) and adversarial~(OOD) samples, by capturing unique batch statistics of these sets~\cite{xie2019intriguing, wang2022removing}. This ``two-domain hypothesis" suggests that employing these batch statistics may enhance the detection of covariate shifts. However, BN can increase adversarial vulnerabilities by encouraging dependence on non-robust features~\cite{benz2021batch}. To address this, a patch-based strategy can be employed, where training with image patches encourages the model to focus on robust features. Consequently, inference with batches of patches from the same scan ensures that batch statistics are derived from a consistent underlying distribution.

AdverX-Ray employs an Adversarial VAE designed to detect OOD samples in the medical imaging domain, particularly from X-ray images. It leverages the discriminator's ability to detect the distribution shifts in adversarially generated content, enabling the system to distinguish between images with different settings from the same machine and those from other machines. This capability is crucial because models trained on images from one machine often fail in quality prediction when using images from other machines due to subtle differences. AdverX-Ray ensures the integrity of medical images, supporting robust and reliable diagnostic systems. The main contributions of this work are listed as follows. (1)~The extension of the DisCoPatch~\cite{caetano2025discopatch} framework into AdverX-Ray, a method tailored for X-ray scans that leverages a patch-based strategy to enhance robustness and efficiency in OOD covariate shift detection. (2)~Publicly made available novel dataset, which is specifically curated for covariate shift detection in X-ray scans. (3)~Comprehensive benchmarking of AdverX-Ray, demonstrating its superior performance and efficiency on newly released data and the BIMCV-COVID19+ dataset.

\section{METHODOLOGY}

\subsection{Datasets}
\label{sec: datasets}
Medical X-ray images, pivotal for diagnostic purposes, contain covariate factors that can compromise image quality and diagnostic accuracy~\cite{manson2019image}. We aim to develop a method to detect changes in the imaging system (or faulty behavior), representing themselves as OOD covariate shifts in the images.

\begin{figure}[!t]
    \centering
    \includegraphics[width=\linewidth]{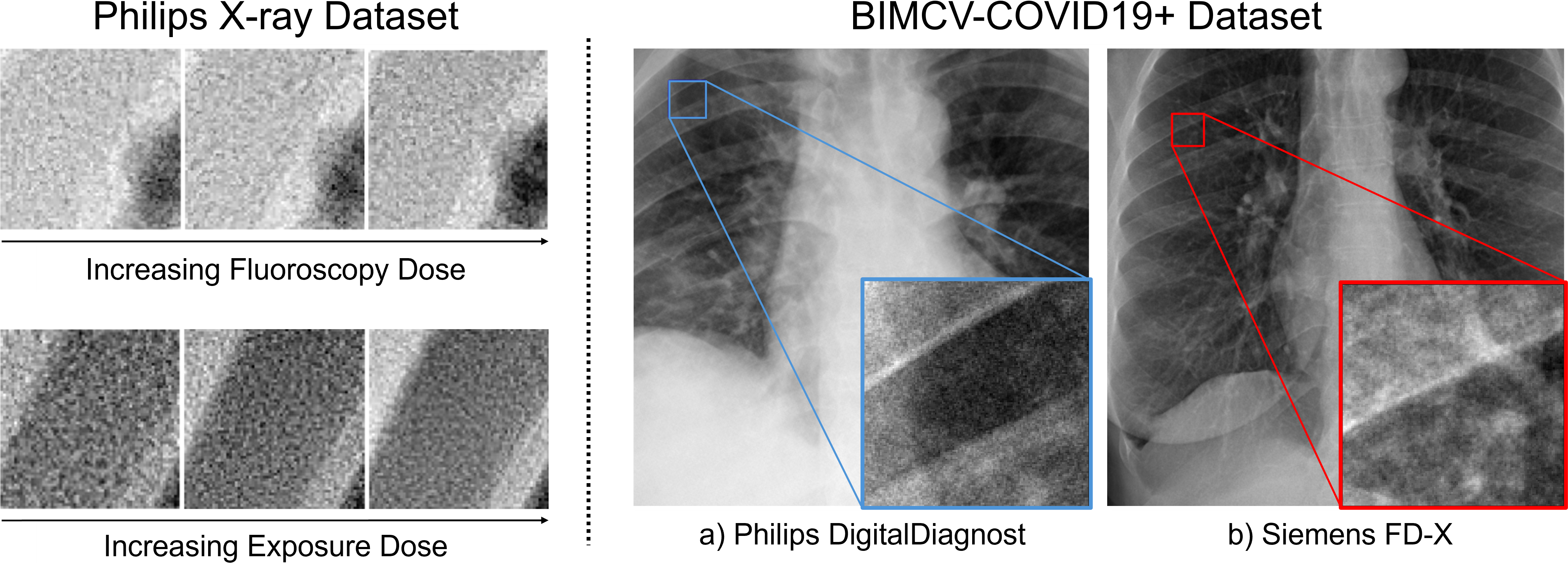}
    \caption{Sample images of the applied datasets. At the left, patches from the Philips X-ray dataset, showcasing different fluoroscopy and exposure doses. At the right, example chest X-ray images and patches from the BIMCV-COVID19+ data, each from a different machine (bottom right shows magnified views).}
    \label{fig:philips_data}
\end{figure}

\textbf{A. Philips X-ray Dataset:} Since we do not have access to such OOD artifacts or faulty images, we capture \textbf{real covariate shifts} in the data by altering the imaging settings. If we can discern the subtle variations caused by changed imaging settings, we hypothesize that the proposed models can measure faulty OOD covariate shifts as an effect of system errors. To this end, we acquire a new dataset of X-ray images containing a standard test object~(clock), using an Azurion Image-Guided Therapy~(IGT) system. We capture images in Dicom format at 12~bits/pixel for different imaging settings, encompassing distinct dose levels using both pulsed fluoroscopy and full radiation modes. These dose variations are operated at varying source-image distances~(SIDs). We organize the dataset in 6 different modes, as follows: Mode~0~(exposure with a normal dose at 110-cm SID), Mode~1~(exposure with a low dose at 110-cm SID), Mode~2~(exposure with a normal dose at 90-cm SID), Mode~3~(exposure with a low dose at 90-cm SID), Mode~4~(fluoroscopy with a normal dose at 110-cm SID), and Mode~5~(fluoroscopy with a low dose at 90-cm SID). Assuming Mode~0 as ID, a progressive covariate shift is expected from Mode~0 to Mode~5 in orders of magnitude, with Mode~5 being the most OOD. On the left side of Figure~\ref{fig:philips_data}, we can observe how different fluoroscopy and exposure doses affect the high-frequency components of the scans. The full dataset consists of 18~Modes and two environments. This dataset will be made publicly available alongside this paper\footnote{\url{https://zenodo.org/records/13924900}}.

\textbf{B. BIMCV-COVID19+ Dataset:} 
This is a large dataset that includes CR, DX, and CT images of COVID-19 patients, as well as their radiographic findings, pathologies, polymerase~(PCR), antibody tests, and radiographic reports from the Medical Image Data Bank of the Valencian Community~(BIMCV). The first iteration of the database includes 1,380~CX, 885~DX, and 163~CT studies from 1,311~COVID-19+ patients, resulting in 3,141~X-ray images and 2,239~CT images. The conducted study concentrates solely on the X-ray scans, clustered into 29~machine models. This grouping simulates real-world scenarios where machine learning models, trained on images from one specific X-ray machine, may need to process images from different machines, potentially leading to performance degradation.  Figure~\ref{fig:philips_data} illustrates the differences between scans from different machines. For our experiments, we define five distinct ID sets, each consisting of images from the five machines with the most scans. The details of the ID sets are provided in Table~\ref{tab:dataset_split}. Each set is divided into 70\% for training and 30\% for testing. 

\begin{table}[h]
    \centering
    \caption{Five different in-distribution sets, which are subsets of the BIMCV-COVID19+ dataset.\label{tab:dataset_split}}
    \vspace{0.2em}
    {\resizebox{0.85\linewidth}{!}{
    \begin{tabular}{l|ccccc}
        \toprule
        \textbf{Manufacturer} & Siemens & Konica & Philips & GE & GMM\\
         & & Minolta & Medical Systems & Healthcare & \\
        \midrule
        \textbf{Model} & FD-X & 0862 & DigitalDiagnost & Thunder Platform & Accord DR \\
        \midrule
        \textbf{\#Images} & 507 & 418 & 357 & 306 & 276 \\
        \bottomrule
    \end{tabular}}}
\end{table}

\subsection{AdverX-Ray}

\begin{figure}[t]
    \centering
    \includegraphics[width=\linewidth]{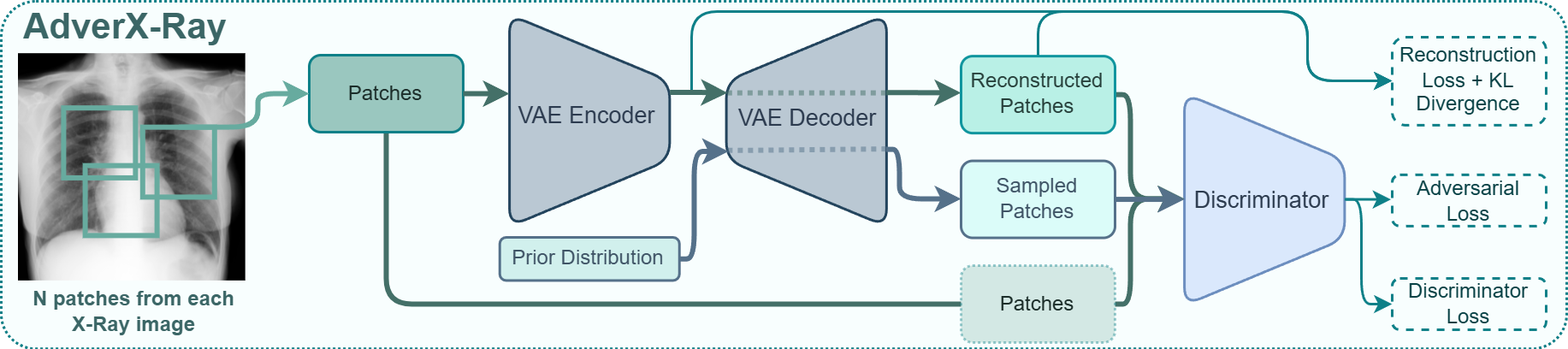}
    \caption{Overview of AdverX-Ray's architecture. During inference, only the discriminator is used.}
    \label{fig:adverxray}
\end{figure}

A common generative-based approach to OOD detection involves using a trained model to identify new unseen samples. Similarly, an adversarially trained discriminator can provide a boundary for the ID set by assessing the probability of a sample being real~(ID) or synthetic~(OOD). We can detect OOD samples by adjusting where the discriminator learns to draw its boundary. It is on this premise that we propose \emph{AdverX-Ray}, shown in Figure~\ref{fig:adverxray}, an X-ray-tailored Adversarial VAE framework that, when fed with a batch of patches from an X-ray scan, can detect covariate shifts effectively. The model is a modified version of the Discriminative Covariate Shift Patch-based Network~(DisCoPatch)~\cite{caetano2025discopatch}. 

\textbf{A. Overview:} The AdverX-Ray approach integrates generative and reconstruction-based strategies to distill information in an unsupervised manner about the ID set and the OOD set boundaries for the discriminator. Unlike conventional adversarial methods, AdverX-Ray leverages the generator output to refine the discriminator capabilities. The VAE is trained to minimize the standard Evidence Lower Bound loss~(ELBO), while producing samples and reconstructions that can also deceive the discriminator. The discriminator is trained to distinguish not only generated and real images, like in a standard Generative Adversarial Network~(GAN), but to differentiate reconstructed images also. Reconstructions from VAEs typically lack high-frequency content~\cite{lin2023catch}, which is important for detecting certain covariate shifts, such as blurriness. Conversely, images generated by GANs often show pronounced high-frequency differences, that may guide the discriminator towards these components too much, thereby neglecting low-frequency elements sometimes~\cite{li2023exploring}. By training the discriminator on both reconstructions and sampled images, and encouraging these to appear realistic, AdverX-Ray tightens the discriminator's understanding of the ID frequency spectrum. The proposed approach leverages the observation that BN can facilitate an adversarially trained discriminator to distinguish individual underlying data distributions by identifying that clean and adversarial images originate from separate domains. By fine-tuning the ID set boundary of the discriminator, we can develop an OOD detection mechanism. Furthermore, by relying solely on batch statistics in the \texttt{BatchNorm} layers of the discriminator, we capture correlations across patches from different zones within each scan, effectively modeling the global structure and relevant features of the X-ray image. However, this approach requires individual processing of scans, yet remains highly efficient due to the lightweight architecture of the discriminator. 

\textbf{B. Patching Strategy:} The proposed patching strategy addresses the issue of black margins in chest X-rays, especially above the shoulders and in areas where black bars are used to maintain the aspect ratio. To address this, we define a fixed region of interest that excludes the top and bottom 20\% of the image, as well as the outermost 20\% both at the left and right sides. The extracted patches have a resolution of 128$\times$128. The image-level score during inference is calculated as the mean of the scores from all patches in that image. 

\subsection{Benchmark}

\textbf{A. Datasets:} For evaluating the Philips dataset, we utilize the entire test set to measure the performance of the models. However, the BIMCV-COVID19+ dataset involves a more nuanced approach. For each subset within this dataset, the test set comprises the ID test split, as defined in Section~\ref{sec: datasets}, and OOD images. The OOD images include the test splits of all other potential ID machines and the images from the remaining 24~machines. To ensure a balanced comparison between ID and OOD images, we perform random image sampling from the OOD set 10~times. The final performance metrics for each model are averaged across these 10~iterations.

\textbf{B. Models:} For the initial evaluation on the internal Philips X-Ray Dataset, we select a state-of-the-art reconstruction-based method, \textit{DDPM-OOD}~\cite{graham2023denoising}, and an explicit density model known for its high-frequency detection capabilities, \textit{GLOW}~\cite{kingma2018glow}. Additionally, we evaluate a conventional \textit{VAE}~\cite{kingma2013auto}, the proposed \textit{AdverX-Ray}, and the \textit{Adversarial VAE}, which serves as the generator component of AdverX-Ray. Following this, we identify the top three approaches and retrain them on the BIMCV-COVID19+ dataset to further validate their performance.

\textbf{C. Performance metrics:} Two metrics are used to evaluate the models, the Area Under the Receiver Operating Characteristic curve~(AUROC) and the False Positive Rate at 95\% True Positive Rate~(FPR95).

\section{RESULTS AND DISCUSSION}

This section presents a comprehensive analysis of the performance achieved by AdverX-Ray across both datasets. We further investigate the influence of varying patch counts on AdverX-Rays's effectiveness, and we evaluate the computational demands associated with the proposed approach.

\subsection{Philips X-ray Dataset}

The results in Table~\ref{tab:aucresults_xray_simple} showcase the performance of the OOD detection methods in the X-ray setting with varying acquisition parameters, which can influence heteroscedastic noise in the signal. Covariate shifts in the X-ray setting are intractable and usually not visible to the non-specialist. Regardless of this and except for the GLOW model trained on typicality~\cite{chali2023typicality}, all methods can detect a shift in the high-level image statistics.

\begin{table*}[!ht]
\centering
\caption{AUROC scores of various methods on detecting OOD covariate shift on the X-Ray dataset. \label{tab:aucresults_xray_simple} }
\vspace*{1mm}
{\resizebox{1.0\linewidth}{!}{
\begin{tabular}{l|ccccc|c} 
\toprule
 \multicolumn{1}{c|}{\underline{ \textbf{X-Ray Mode 0 ID} } }           &                            \multicolumn{5}{c|}{\underline{\textbf{X-Ray Other OOD} }} & \textbf{Average} \\
 Method \enspace &\enspace Mode 1 \enspace &\enspace  Mode 2 \enspace &\enspace Mode 3 \enspace &\enspace  Mode 4 \enspace &\enspace Mode 5 & \enspace AUROC$\uparrow$\slash FPR95$\downarrow$ \\
\midrule
DDPM~\cite{graham2023denoising} (T250: LPIPS) & 86.4 & 88.5 & 84.2 & 89.3& 96.5 & 89.0\slash69.4\\
DDPM~\cite{graham2023denoising} (T250: LPIPS + MSE) & 79.9& 83.0 & 81.1 & 89.0 & 95.8 & 85.8\slash67.3\\
\midrule
VAE~\cite{kingma2013auto} (ELBO) & 69.7& 96.1 & 88.9 & 91.3 & 98.1 & 88.8\slash25.0\\
AVAE~\cite{plumerault2021avae} (ELBO + Adv Loss) & 65.5 &  94.8 & 86.5& 89.5 & 97.2 & 86.7\slash28.9\\
GLOW~\cite{kingma2018glow} (log-likelihood) & 89.3 & 98.8 & 99.8 & 99.9 & \textbf{100.0} & 97.6\slash 8.7\\
GLOW~\cite{chali2023typicality} (Typicality) & 30.4& 16.5 & 15.1 & 10.4& 11.1 & 16.7\slash99.8 \\
AdverX-Ray (proposed) & \textbf{99.5} & \textbf{99.9}  & \textbf{99.9} & \textbf{100.0} & \textbf{100.0} & \textbf{100.0\slash0.0}  \\
\bottomrule 
\end{tabular}}}
\end{table*}

\subsection{BIMCV-COVID19+ Dataset}

Table~\ref{tab:aucresults_bimcv_simple} demonstrates the AdverX-Ray model prediction quality for all ID sets, compared to the VAE and GLOW. GLOW achieves poor results because of a known phenomenon in which it attributes higher likelihoods to OOD images\cite{normalizingfail}. Interestingly, this limitation is inconsistent, occurring only for certain machines. AdverX-Ray convincingly surpasses the VAE and GLOW in all test sets, except for the GE machine, which achieves close performance to the VAE.

\begin{table*}[!h]
\centering
\caption{AUROC scores of various methods on detecting OOD covariate shift on the BIMCV-COVID19+ dataset. \label{tab:aucresults_bimcv_simple} }
\vspace*{1mm}
{\resizebox{1.0\linewidth}{!}{
\begin{tabular}{l|ccccc|c} 
\toprule
 \multirow{2}{*}{\textbf{Method}} & \multicolumn{5}{c|}{\underline{\textbf{ID Machine}}} & \textbf{Average} \\
\enspace &\enspace Siemens \enspace &\enspace Konica \enspace &\enspace Philips \enspace &\enspace GE \enspace &\enspace GMM \enspace & \enspace AUROC$\uparrow$\slash FPR95$\downarrow$\\
 \midrule
VAE~\cite{kingma2013auto} (ELBO) & 49.7 & 30.6 & 38.8 & \textbf{98.6} & 62.3 & 56.0\slash81.5\\
GLOW~\cite{kingma2018glow} (log-likelihood) & 31.4 & 34.2 & 18.9 & 88.1 & 90.6 & 52.7\slash95.2\\
AdverX-Ray (proposed) & \textbf{96.3}  & \textbf{96.3}  & \textbf{96.1}  & 92.4  & \textbf{99.7}  & \textbf{96.2/19.7} \\ 
\bottomrule 
\end{tabular}}}
\end{table*}

\begin{figure}[!h]
    \centering
    \includegraphics[width=0.9\linewidth]{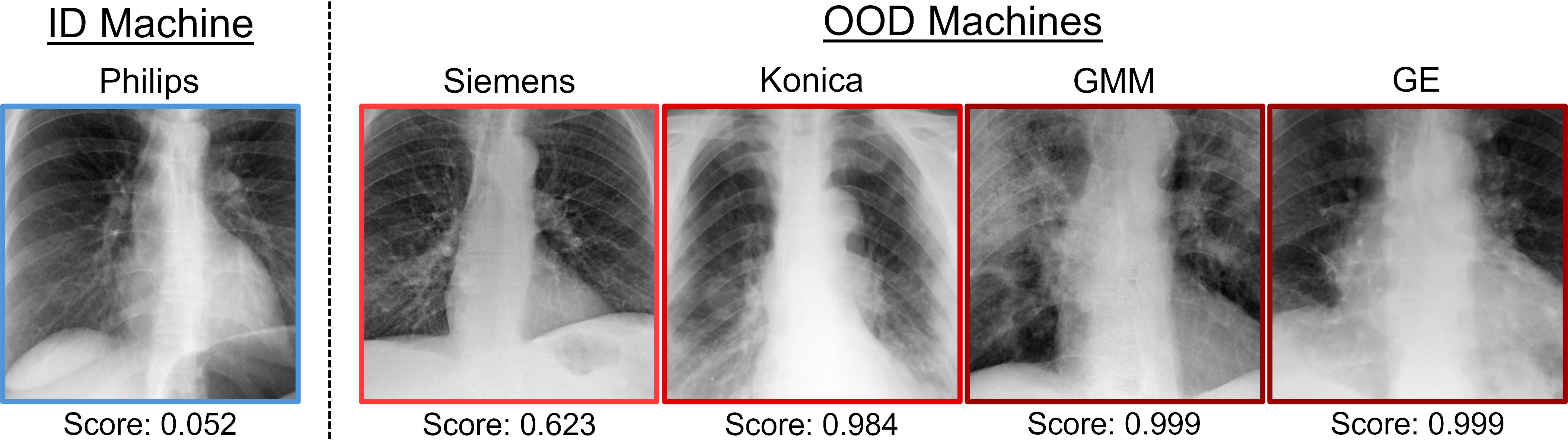}
    \caption{Examples of OOD scores generated by the AdverX-Ray model trained on the Philips X-ray scans.}
    \label{fig:scores}
\end{figure}

Although Table~\ref{tab:aucresults_xray_simple} indicates that the VAE and GLOW models effectively identify high-frequency covariate shifts within the Philips X-Ray dataset, their sole focus on these components is inadequate to identify scans from alternative manufacturers. AdverX-Ray's broader frequency coverage significantly overcomes this constraint. It should be noted that AdverX-Ray achieves these results with a 15-MB model, compared to the VAE's 126~MB and GLOW's 120~MB. These results are obtained with 64~random patches per image and correspond to the average of 5~execution cycles. Figure~\ref{fig:scores} depicts the OOD scores generated by the AdverX-Ray model trained on scans from the Philips machine; the model consistently attributes higher OOD scores for scans from other manufacturers.

\subsection{Impact of the number of patches}

\begin{figure}[!h]
    \centering
    \includegraphics[width=\linewidth]{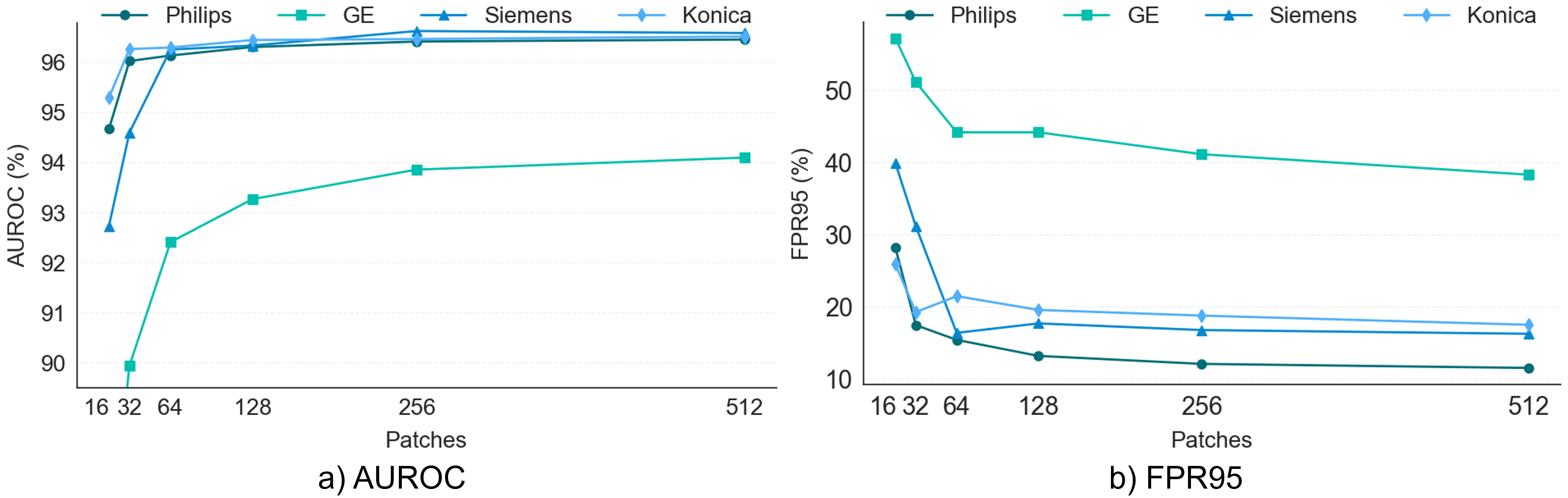}
    \caption{AUROC and FPR scores for AdverX-Ray on the BIMCV-COVID19+ dataset for varying patch counts per image. GMM scans are not included due to the low degree of variability for different patch counts.}
    \label{fig:aurocfpr}
\end{figure}

The results in Table~\ref{tab:aucresults_adverx} validate the prior choice of using 64~patches per image, as this configuration achieves a strong balance between performance and computational efficiency. Although increasing the patch count per image does yield incremental improvements, particularly in FPR95, the gains are marginal beyond 64~patches. This effect is more noticeable in Figure~\ref{fig:aurocfpr}, in which it is possible to observe that the major improvements for a larger number of patches occur when using the GE machine, in which the model achieves its lowest performance.

\begin{table}[!h]
    \centering
    \caption{AUROC scores for AdverX-Ray on the BIMCV-COVID19+ dataset with varying patch counts per image. \label{tab:aucresults_adverx} }
    \vspace*{1mm}
{\resizebox{1.0\linewidth}{!}{
    \begin{tabular}{@{\extracolsep{12pt}}l|ccccc|c@{}}
\toprule
 \textbf{No. of} & \multicolumn{5}{c|}{\underline{\textbf{ID Machine}}} &\textbf{Average} \\
\textbf{Patches} &\enspace Siemens \enspace &\enspace Konica \enspace &\enspace Philips \enspace &\enspace GE \enspace &\enspace GMM \enspace & \enspace AUROC$\uparrow$\slash FPR95$\downarrow$\\
\midrule
        \hspace{2mm}16 & 92.7  & 95.3  & 94.7  & 86.0  & 99.5  & 93.6/30.6  \\ 
        \hspace{2mm}32 & 94.6  & 96.3  & 96.0  & 89.9  & 99.5  & 95.3/24.2  \\ 
        \hspace{2mm}64 & 96.3  & 96.3  & 96.1  & 92.4  & 99.7  & 96.2/19.7  \\ 
        \hspace{2mm}128 & 96.3  & 96.5  & 96.3  & 93.3  & 99.7  & 96.4/19.1  \\ 
        \hspace{2mm}256 & 96.6  & 96.5  & 96.4  & 93.9  & 99.7  & 96.6/18.0  \\ 
        \hspace{2mm}512 & 96.6  & 96.5  & 96.5  & 94.1  & 99.7  & 96.7/16.9  \\
        \bottomrule
    \end{tabular}}}
\end{table}

\subsection{Computational Requirements}

To assess the computational requirements, Table~\ref{tab:comp_resources} provides the model size, total number of parameters, average training time for all BIMCV-COVID19+ ID sets, and average inference time to obtain the score for a single image. All models are evaluated with 64 patches per image. AdverX-Ray is the smallest and fastest model and requires significantly less time than GLOW for training. 

\begin{table}[!ht]
\centering
\caption{Computational resources required for training and deployment of the models (smallest in bold). \label{tab:comp_resources} }
\vspace*{1mm}
{\resizebox{1.0\linewidth}{!}{
\begin{tabular}{l|cccc} 
\toprule
\textbf{Method} \enspace &\enspace \textbf{Size (MB)} \enspace&\enspace \textbf{\#Parameters} \enspace&\enspace \textbf{Training Time (s)} \enspace&\enspace \textbf{Inference Speed (s)} \\
\midrule
VAE~\cite{kingma2013auto} (ELBO) & 126 & 33,036,737 & \textbf{1,480} & 0.005\\
GLOW~\cite{kingma2018glow} (LL) & 120 & 31,570,304 & 28,482 & 2.369\\
AdverX-Ray (Proposed) & \textbf{15} & \textbf{3,928,962}  & 4,162 & \textbf{0.003}\\ 
\bottomrule 
\end{tabular}}}
\end{table}

\section{CONCLUSION} 

This paper introduces a novel image quality evaluation model, AdverX-Ray, an Adversarial VAE framework tailored to detecting OOD covariate shift in X-ray imaging. AdverX-Ray combines reconstructed and generated image patches in its training process to carefully distinguish various frequency-spectrum perturbations. By showcasing its superior performance in detecting changes in imaging settings~(Philips X-ray) and images from different scanners~(BIMCV-COVID19+), the model ensures the integrity of medical imaging systems and the potential CAD methods applied to the resulting images. AdverX-Ray’s efficient and lightweight architecture makes it suitable for real-world applications requiring minimal computation resources and offers reliable deployment in clinical settings.

\acknowledgments 
 
We would like to acknowledge the Philips IGT Test Automation team for their invaluable assistance with data collection. This research was funded by the European Xecs Eureka TASTI Project. This work used the Dutch national e-infrastructure with the support of the SURF Cooperative using grant no. EINF-10021.

\bibliography{report} 
\bibliographystyle{spiebib} 

\end{document}